\documentclass[11pt,a4paper]{article}
\usepackage[hyperref]{acl2021}
\usepackage{times}
\usepackage{latexsym}

\usepackage{times}
\usepackage{latexsym}
\usepackage{graphicx}
\usepackage{url}
\usepackage{amsmath}
\usepackage{breqn}
\usepackage{makecell}
\usepackage{tabularx}
\usepackage{bm}

\newcommand*{\affaddr}[1]{#1}
\newcommand*{\affmark}[1][*]{\textsuperscript{#1}}

\usepackage{microtype}

\newcommand{\IntroductionMotivatingExample}{
\begin{table}[t!]
\centering
\small
\begin{tabularx}{\columnwidth}{|X|}
\hline
\textbf{Headline:} SuperBowl
 \\ \hline
\textbf{Snippet:} Whether you’re a football fan or not, what do you like about Super Bowl Sunday?
 \\ \hline
\textbf{Comment:} ... In my opinion I think the Falcons will stomp the patriots. I think Tom Brady will choke the Super Bowl. ...
\\ \hline
\textbf{Comment:}  I am big Arizona Cardinals fan so when they didn't even make the playoffs i was upset. ...
\\ \hline
\textbf{Comment:} I'm not a very big football fan at all. So when it comes to Superbowl Sunday, I'm in it for the commercials and the half time show. ...
\\ \hline
\textbf{Comment:}  I am not exactly a football fan, but I enjoy watching the Super Bowl....
\\ \hline
\textbf{...}
\\ \hline
\textbf{Summary:} \\ 
Several commenters list their favorite things about the Super Bowl, including half-time shows, the funny commercials, the Puppy Bowl, eating food, and spending time with family. A couple of commenters admit to not being football fans but still enjoying the Super Bowl. Some commenters discuss whether they thought the Falcons or the Patriots were going to win, while others list teams they wish were in the game.
 \\
\hline
\end{tabularx}
\caption{Example summary of comments from a New York Times article discussing people's favorite parts of the Super Bowl. The summary is an analysis of the comments and quantifies the viewpoints present.}
\label{tab:example}
\end{table}
}

\newcommand{\ArgumentGraph}{
\begin{figure*}[t!]
    \centering
    \includegraphics[scale=0.40]{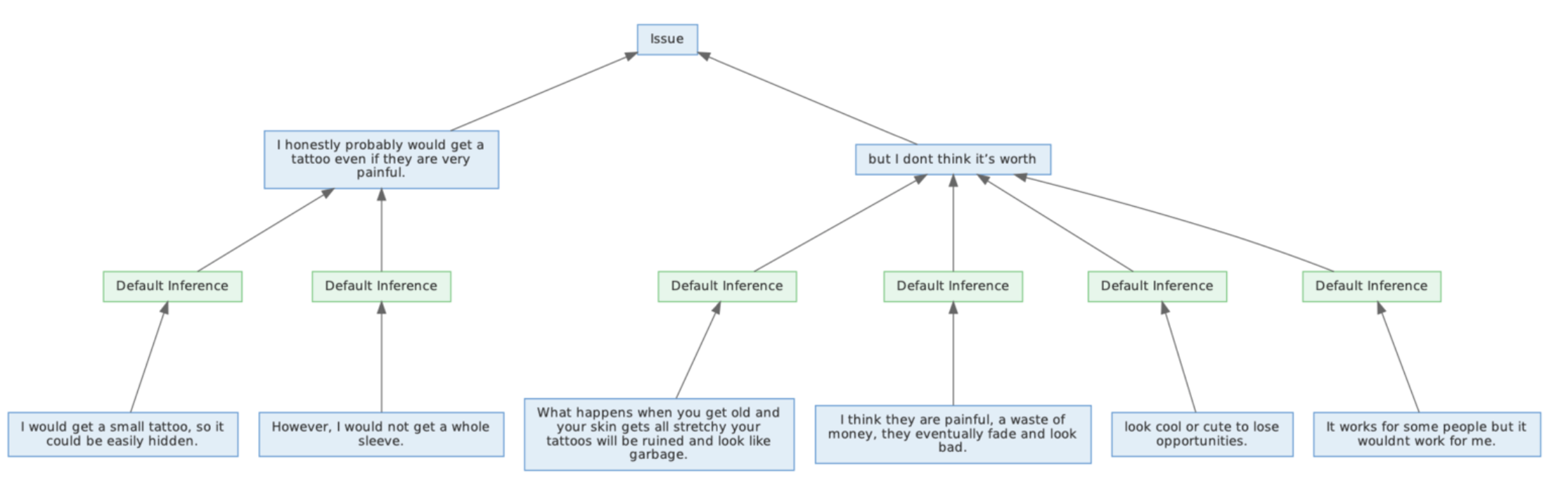}
    \caption{Sample argument subgraph construct from NYT news comments illustrating varying viewpoints. Claims ``I honestly..." and ``but I dont.." are entailed by premises, connected through \texttt{Default Inference} nodes, and opposing claims are connected through \texttt{Issue} nodes.
    }
    \label{fig:argument_graph}
\end{figure*} 
}

\newcommand{\TableHumanEvaluation}{
\begin{table}[t!]
\centering
\resizebox{\columnwidth}{!}{\begin{tabular}{ c  c c   c c } \Xhline{2\arrayrulewidth}
            \textbf{Target Dataset} &  \multicolumn{2}{c}{BART} &  \multicolumn{2}{c}{BART-arg}  \\
              &  Relevance & Consistency & Relevance & Consistency \\
          \Xhline{2\arrayrulewidth} 
          Reddit   &  3.39 (0.13) & 3.40 (0.12) &  3.47 (0.12) &  3.41 (0.10) \\
          AMI   &  4.07 (0.16) & 3.67 (0.16) &  4.13 (0.17) &  3.70 (0.17) \\
            \Xhline{2\arrayrulewidth}
\end{tabular}}
\caption{Mean relevance and factual consistency annotations for BART and BART-arg outputs on Reddit and AMI. Standard errors are reported in parentheses.}
\label{tab:table_human_evaluation}
\end{table}
}

\newcommand{\ExampleSummariesFirst}{
\begin{table}[t!]
\centering
\small
\begin{tabularx}{\columnwidth}{|X|}
\hline
\textbf{Title}: What should I draw next?   \textbf{Subreddit}: pokemon \\ \hline
 A FB user said they wanted to see a Zapdos, but depending on what people say on here, if any, I may do one that you all choose! So help out if you want!  Or don't. :) \\ \hline
\textbf{Source Documents:} \\ \hline
 1.  I'd like to see any of the Legendary Beasts, Raikou, Suicune, Entei. \\
 2.  Ooo I'm kinda thinking of Suicune now. Good choice! I'll consider it for sure! If not, it will definitely be very soon!    \\
3.   Pancham   $\leftarrow$ Oooo good choice. That'll be one of the next ones too.   \\
4.   If you have the time, I'd love to see Reshiram!  $\leftarrow$ I'll definitely do that as one of the next ones!  Thank you very much! \\ 
5.   Seel?  $\leftarrow$ With your username, I may just do that soon.  $\leftarrow$  Cool!    \\
6.   MAGIKARP   $\leftarrow$ Dear Lord. \\
7.  Aron \& lairon!  \\
8.   You should girafarig because it's the best pokemon ever created no matter what ANYONE SAYS.   $\leftarrow$ While I admire your steadfast courage in how awesome Girafig is, I may wait on that one. :P BUT. Maybe. :) \\ \hline
\textbf{BART:} Most commenters say that they would like to see a Zapdos or Suicune. A few other suggestions are Girafig, Arron, lairon, Magikarp, Raikou, Entei, Reshiram, Seel, and Pancham. One commenter says that they want to see Girafarig, but another says that he may wait on that one for a while. \\ \hline
\textbf{BART-arg:} Most commenters say that they would like to see a Zapdos, Raikou, Suicune, Entei, Pancham Aron, and lairon. A few commenters say they would love to see Girafarig. One commenter says that he would like Girafig because it's the best pokemon ever created no matter what anyone says about it. Another commenter says they want to see Reshiram. \\ \hline
 \textbf{Gold Summary:} Commenters suggest ideas of the next Pokemon for the original poster to draw next. Each commenter has a different suggestion including Raikou, Reshiram, Seel, Magikarp and Lairon, and the original poster comments their thoughts on each of the suggestions with either approval or a joking disapproval. \\ \hline
\end{tabularx}
\caption{Example source documents and summaries from Reddit data in which the models are faithful to the source and largely relevant.}
\label{tab:example_summaries_first}
\end{table}
}

\newcommand{\ExampleSummariesSecond}{
\begin{table}[t!]
\centering
\small
\begin{tabularx}{\columnwidth}{|X|}
\hline
\textbf{Title}:  Complete noob here. Any advice welcome!   \textbf{Subreddit}: personalfinance \\ \hline
I've always been responsible when it comes to my money. In the sense that I dont spend what I dont have, I save my money, and I watch what I spend. But now I've recently graduated (with 28k in debt) and want to do better. I've got, what I think, is a decent plan to pay off my debt. ... But what else should I be doing? ... \\ \hline
\textbf{Source Documents:} \\ \hline
1. Please read the information found in the [FAQ](http://www.reddit.com/r/personalfinance/wiki/faq), particularly ...  \\
2. Make sure any extra money is going towards the principal. \\
3. If it makes you feel better there was never a class where this was explained. ... The best way to learn is to check out the FAQs on the sidebar ... \\
4. ... You should also save up some cash in a bank account and consider that your emergency fund,... \\
5. The two common retirement accounts are 401ks and IRAs  \\
6. Invest in pumpkins. \\ \hline
\textbf{BART:} Most commenters agree that the best way to learn about investing is to read the FAQ. A few commenters say that if you have debt, then you should start making contributions to a 401k retirement plan. \textcolor{red}{One commenter says that you need to start a small business to pay off your debt}, and another says that it would be a good idea to have some cash in a bank account. \textcolor{red}{One commentator says that they think the best thing to do is start a business.}  \\ \hline
\textbf{BART-arg:} Most commenters agree that the best way to start a small business is to pay off your debt. \textcolor{red}{Some commenters} say that you should also invest in pumpkins. One commenter says that if you don’t have a 401k, then you should start making contributions to a Roth retirement plan. Another commenter suggests that you invest in a bank account to build up an emergency fund. One commentator says that you need to be responsible when it comes to your money. \\ \hline
\textbf{Gold Summary:} Commenters offer their advice for settling a debt. Several commenters agree that utilizing the avalanche model is the correct way to pay the debt. Another also suggests having an emergency fund for any unexpected bills or necessary expenses. In understanding finance, commenters stress the need to check the sidebar for FAQs.  \\ \hline
\end{tabularx}
\caption{Example source documents and summaries from Reddit data in which the models fail to completely capture salience while remaining faithful to the input.}
\label{tab:example_summaries_second}
\end{table}
}

\newcommand{\BARTTransferFullResults}{
 \begin{table}[h!]
\resizebox{\columnwidth}{!}{\begin{tabular}{ c   c  c  c  }
\Xhline{2\arrayrulewidth}
             \textbf{Data/Method}  &  \textbf{BART} & \textbf{BART-arg-graph} & \textbf{BART-arg-filtered}\\  \hline
            NYT &  35.91/9.22/31.28 & 36.02/9.60/32.34 & \textbf{36.60}/\textbf{9.83}/\textbf{32.61} \\ \hline  
            Reddit &  35.50/10.64/32.57 & 36.39/\textbf{11.38}/\textbf{33.57} &  \textbf{36.51}/11.02/33.14 \\ \hline  
            Stack &  39.61/10.98/35.35 & \textbf{39.73}/\textbf{11.17}/\textbf{35.52} & 39.40/10.98/35.51 \\ \hline  
            Email &  \textbf{41.46}/\textbf{13.76}/ \textbf{37.70} & 39.05/12.14/35.99 &  40.32/12.97/36.90  \\ \hline  
            \Xhline{2\arrayrulewidth}
\end{tabular}}
\caption{Full ROUGE-1/2/L results for vanilla BART, -arg-graph, and -arg-filtered input. All are trained on 200 points from ConvoSumm.}
\label{tab:bart_transfer_full}
\end{table}
}

\newcommand{\DatasetComparison}{
 \begin{table*}[t!]
\resizebox{\textwidth}{!}{\begin{tabular}{c c c c c c}
\Xhline{2\arrayrulewidth}
             \textbf{Dataset} & \textbf{\% novel n-grams} & \textbf{Extractive Oracle}  &  \textbf{Summary Length} & \textbf{Input Length}  & \textbf{\# Docs/Example} \\  \hline
            NYT & 36.11/79.72/94.52 & 36.26/10.21/31.23 & 79 & 1624 & 16.95  \\ \hline  
            Reddit & 43.84/84.98/95.65 & 35.74/10.45/30.74  & 65 & 641 & 7.88 \\ \hline  
            Stack & 35.12/77.91/93.56 &  37.30/10.70/31.93  & 73 & 1207 & 9.72   \\ \hline  
            Email & 42.09/83.27/93.98 & 40.98/15.50/35.22  & 74 & 917 & 4.95  \\ \hline  
            \Xhline{2\arrayrulewidth}
\end{tabular}}
\caption{Statistics across dataset sources in ConvoSumm, showing novel uni/bi/tri-grams, ROUGE-1/2/L extractive oracle scores, the average input and summary lengths (number of tokens), as well as the number of documents per example, where each comment/post/answer/email is considered a document.}
\label{tab:statistics}
\end{table*}
}

\newcommand{\DatasetComparisonMDS}{
 \begin{table}[t]
\resizebox{\columnwidth}{!}{\begin{tabular}{c c c c c c}
\Xhline{2\arrayrulewidth}
             \textbf{Dataset/Method} & \textbf{Inter-document Similarity} & \textbf{Redundancy} & \textbf{Layout Bias} \\  \hline
            NYT & -11.71 & -0.23 & 0.2/0.5/0.3  \\ \hline  
            Reddit & -7.56 & -0.49 & 0.2/0.5/0.2  \\ \hline  
            Stack & -9.59 & -0.27 & 0.2/0.3/0.4 \\ \hline  
            Email & -1.76 & -0.18 & 0.3/0.4/0.3 \\ \hline  
            \Xhline{2\arrayrulewidth}
\end{tabular}}
\caption{Multi-document summarization-specific dataset analysis on our proposed datasets with metrics introduced in \citet{dey2020corpora}:   inter-document similarity (father from zero is less similarity), redundancy (father from zero is less overall redundancy of semantic units), and start/middle/end layout bias.}
\label{tab:statistics_mds}
\end{table}
}

\newcommand{\BaselineResults}{
 \begin{table}[t]
\resizebox{\columnwidth}{!}{\begin{tabular}{c c c c}
\Xhline{2\arrayrulewidth}
             \textbf{Dataset/Method} & \textbf{Lexrank} &  \textbf{Textrank} & \textbf{BERT-ext} \\  \hline
            NYT & 22.30/3.87/19.14 & 25.11/3.75/20.61 & 25.88/3.81/22.00  \\ \hline  
            Reddit & 22.71/4.52/19.38 & 24.38/4.54/19.84 & 24.51/4.18/20.95 \\ \hline  
            Stack & 26.30/5.62/22.27 & 25.43/4.40/20.58 & 26.84/4.63/22.85 \\ \hline  
            Email & 16.04/3.68/13.38 & 19.50/3.90/16.18 & 25.46/6.17/21.73 \\ \hline  
            \Xhline{2\arrayrulewidth}
\end{tabular}}
\caption{ROUGE-1/2/L results for extractive LexRank \cite{erkan2004lexrank}, TextRank \cite{mihalcea2004textrank}, and BERT-based \cite{miller2019leveraging} models.}
\label{tab:extractive_baselines}
\end{table}
}

\newcommand{\BARTTransferResults}{
 \begin{table}[t]
\resizebox{\columnwidth}{!}{\begin{tabular}{c c c}
\Xhline{2\arrayrulewidth}
             \textbf{Data/Method}  &  \textbf{BART} & \textbf{BART-arg} \\  \hline
            NYT &  35.91/9.22/31.28 & \textbf{36.60}/\textbf{9.83}/\textbf{32.61} \\ \hline  
            Reddit &  35.50/10.64/32.57 & \textbf{36.39}/\textbf{11.38}/\textbf{33.57}  \\ \hline  
            Stack &  39.61/10.98/35.35 & \textbf{39.73}/\textbf{11.17}/\textbf{35.52}  \\ \hline  
            Email &  \textbf{41.46}/\textbf{13.76}/\textbf{37.70} &  40.32/12.97/36.90  \\ \hline  
            \Xhline{2\arrayrulewidth}
\end{tabular}}
\caption{ROUGE-1/2/L results for vanilla BART as well as one trained on argument-mining input. Both are trained on 200 points from ConvoSumm.}
\label{tab:bart_transfer}
\end{table}
}

\newcommand{\BARTMeetingResults}{
 \begin{table}[t]
\resizebox{\columnwidth}{!}{\begin{tabular}{c c c}
\Xhline{2\arrayrulewidth}
             \textbf{Method/Dataset} & \textbf{AMI} &  \textbf{ICSI}  \\ \hline
            HMNet & 53.02/18.57/- & \textbf{46.28}/10.60/- \\ \hline  
            DDA-GCN & 53.15/\textbf{22.32}/- & - \\ \hline  
            Longformer-BART & 54.20/20.72/51.36 & 43.03/\textbf{12.14}/40.26 \\ \hline
           Longformer-BART-arg & \textbf{54.47}/20.83/\textbf{51.74} & 44.17/11.69/\textbf{41.33} \\ \hline  
            \Xhline{2\arrayrulewidth}
\end{tabular}}
\caption{ROUGE-1/2/L results for DDA-GCN \cite{feng2020dialogue} and HMNet \cite{zhu2020a} on the AMI and ICSI meeting summarization dataset along with our Longformer and Longformer-arg models.}
\label{tab:meeting}
\end{table}
}

\newcommand{\BARTBenchmarkOthers}{
 \begin{table}[t]
\resizebox{\columnwidth}{!}{\begin{tabular}{c c c}
\Xhline{2\arrayrulewidth}
             \textbf{Dataset/Method} & Our results  &  Previous SOTA \\  \hline
            SAMSum & \textbf{52.27/27.82/47.92} & 49.30/25.60/47.70   \\ \hline   
            CQASUMM  & \textbf{32.79/6.68/28.83} & 31.00/5.00/15.20   \\ \hline  
            BC3 & 39.59/13.98/21.20 & -   \\ \hline  
            ADS & 37.18/11.42/21.27 & -   \\ \hline  
            SENSEI & 34.57/7.08/16.80 & -   \\ \hline  
            \Xhline{2\arrayrulewidth}
\end{tabular}}
\caption{Benchmarking results on conversational datasets such as SAMSum \cite{gliwa-etal-2019-samsum} and CQASUMM \cite{chowdhury2018cqasumm} and initial neural abstractive summarization results for email (BC3) \cite{JanAAAI08}, debate discussion forums (ADS) \cite{misra-etal-2015-argumentative-dialogue-corpus}, and news comments (SENSEI) \cite{barker2016sensei}.}
\label{tab:benchmark_others}
\end{table}
}

\aclfinalcopy 

\title{ConvoSumm: Conversation Summarization Benchmark \\ and Improved Abstractive Summarization with Argument Mining}

\author{
 \textbf{Alexander R. Fabbri}\affmark[$\dagger$]  \quad \textbf{Faiaz Rahman}\affmark[$\dagger$]  \quad \textbf{Imad Rizvi}\affmark[$\dagger$]  \quad \textbf{Borui Wang}\affmark[$\dagger$] \\
  \quad \textbf{Haoran Li }\affmark[$\ddagger$] 
  \quad \textbf{Yashar Mehdad}\affmark[$\ddagger$] \quad \textbf{Dragomir Radev}\affmark[$\dagger$] \\
\affaddr{\affmark[$\dagger$] Yale University} 
  \affaddr{\affmark[$\ddagger$] Facebook AI} \\
  \texttt{\{alexander.fabbri, faiaz.rahman, imad.rizvi,}\\ \texttt{borui.wang, dragomir.radev\}@yale.edu} \\
          \texttt{\{aimeeli, mehdad\}@fb.com} 
}

\date{}

\begin{document}
\maketitle


\begin{abstract}
While online conversations can cover a vast amount of information in many different formats, abstractive text summarization has primarily focused on modeling solely news articles. This research gap is due, in part, to the lack of standardized datasets for summarizing online discussions. To address this gap, we design annotation protocols motivated by an issues--viewpoints--assertions framework to crowdsource four new datasets on diverse online conversation forms of news comments, discussion forums, community question answering forums, and email threads. We benchmark state-of-the-art models on our datasets and analyze characteristics associated with the data. To create a comprehensive benchmark, we also evaluate these models on widely-used conversation summarization datasets to establish strong baselines in this domain. Furthermore, we incorporate argument mining through graph construction to directly model the issues, viewpoints, and assertions present in a conversation and filter noisy input, showing comparable or improved results according to automatic and human evaluations. 
\end{abstract}

\section{Introduction}\label{sec:introduction}
Automatic text summarization is the process of outputting the most salient parts of an input in a concise and readable form. Recent work in summarization has made significant progress due to introducing large-scale datasets such as the CNN-DailyMail dataset \cite{nallapati-etal-2016-abstractive} and the New York Times dataset \cite{sandhaus2008new}. Furthermore, the use of large self-supervised pretrained models such as BART \cite{lewis-etal-2020-bart} and Pegasus \cite{zhang2019pegasus} has achieved state-of-the-art performance across summarization tasks and strong performance in zero and few-shot settings \cite{fabbri2020improving}. However, less work has focused on summarizing online conversations. Unlike documents, articles, and scientific papers, which contain specific linguistic structures and conventions such as topic sentences and abstracts, conversational text scatters main points across multiple utterances and between numerous writers. As a result, the text summarization task in the conversational data domain offers a challenging research field to test newly-developed models \cite{chen2020multiview}.
\IntroductionMotivatingExample
\par
Recently, \citet{gliwa2019samsum} introduced a dataset for chat-dialogue conversation summarization consisting of 16k examples, the first large-scale dataset of its kind. Previous work in conversation summarization was limited by the data available and focused primarily on meeting summarization, such as the AMI \citep{kraaij2005ami} and ICSI \citep{janin2003icsi} datasets. The datasets used in recent conversation papers are often not uniform, ranging from visual dialogue data \cite{goo2018abstractive} to customer-service dialogues \cite{yuan2019abstractive}, not initially intended for summarization. The availability of benchmark datasets for comparing methods has limited work in other conversation summarization domains and thus likely inhibited progress \cite{kryscinski-etal-2019-neural, fabbri2020summeval}.
\par
We aim to address this research gap by crowdsourcing a suite of four datasets, which we call \textbf{ConvoSumm}, that can evaluate a model's performance on a broad spectrum of conversation data. In determining the domains of data to collect, we use the general definition of conversation as ``any discourse produced by more than one person'' \cite{ford1991linguistics}. We identify several key categories of data for which standard human-created development and testing datasets do not exist, namely (1) news article comments, (2) discussion forums and debate, (3) community question answering, and (4) email threads. We design annotation protocols motivated by work in quantifying viewpoints present in news comment data \cite{barker2016summarizing} to crowdsource 250 development and 250 test examples for each of the above domains. We provide an example of comments to a New York Times news article, and our crowdsourced summary in Table \ref{tab:example}.
\par
In addition to introducing manually-curated datasets for conversation summarization, we also aim to unify previous work in conversation summarization. Namely, we benchmark a state-of-the-art abstractive model on several conversation datasets: dialogue summarization from SAMSum \citep{gliwa-etal-2019-samsum}, heuristic-generated community question answering from CQASumm \cite{chowdhury2018cqasumm}, meeting summarization data from AMI and ICSI, and smaller test sets in the news comments, discussion forum, and email domains. We believe that such benchmarking will facilitate a more straightforward comparison of conversation summarization models across domains. 
\par
To unify modeling across these conversational domains, we propose to use recent work in end-to-end argument mining \cite{lenz2020towards,stab2014identifying,chakrabarty2020ampersand} to instantiate the theoretical graph framework which motivated our annotation protocol, proposed by \citet{barker2016summarizing} for conversation summarization. This protocol is employed to both identify and use the ``issues--viewpoints--assertions'' argument structure (discussed in Related Work) for summarizing news comments. We construct this argument graph using entailment relations, linearize the graph, train a graph-to-text model \citep{ribeiro2020investigating}, and experiment with argument mining as a way to reduce noise in long-text input. 
\par
Our contributions are the following: (1) we crowdsource datasets for four domains of conversational data and analyze the characteristics of our proposed datasets; (2) we benchmark state-of-the-art models on these datasets as well as previous widely-used conversation summarization datasets to provide a clear baseline for future work; and (3) we apply argument mining to model the structure of our conversational data better as well as reduce noise in long-text input, showing comparable or improved results in both automatic and human evaluations.\footnote{For reproducibility of our findings, we will make our data and code publicly available at \url{https://github.com/Yale-LILY/ConvoSumm}.}
\section{Related Work}\label{sec:related_work}
\paragraph{Modeling Conversation Summarization}
 Early approaches to conversation summarization consisted of feature engineering \citep{xie2008feature}, template selection methods \citep{oya-etal-2014-template}, and statistical machine learning approaches \citep{galley-2006-skip, wang-cardie-2013-domain}. More recent modeling approaches for dialogue summarization have attempted to take advantage of conversation structures found within the data through dialogue act classification \cite{goo-2018-abstractive}, discourse labeling \cite{ganesh-2019-abstractive}, topic segmentation \cite{liu-2019-topic}, and key-point analysis \cite{liu-2019-automatic}. \citet{chen2020multiview} utilize multiple conversational structures from different perspectives in its sequence-to-sequence model. However, such approaches focus exclusively on dialogue summarization, and it is not trivial to extend such methods to longer conversations with many more participants. We thus introduce a method to model the structure of the discourse over the many-party conversation. 
\par
Several existing works have focused on conceptualizing conversation structure for summarization and how to present this structure to end-users. \citet{barker-etal-2016-sensei} propose a conversation overview summary that aims to capture the key argumentative content of a reader comment conversation. \citet{misra2017using} use summarization as a means of probing online debates to discover central propositions, which they cluster to identify argument facets.  \citet{barker-gaizauskas-2016-summarizing} identify three key components of conversational dialogue: \textit{issues} (that individuals discuss), \textit{viewpoints} (that they hold about these issues), and \textit{assertions} (that they make to support their viewpoints). We build on this framework and advances in argument mining for end-to-end training for summarization.
\paragraph{Argument Mining} Work in argument mining \citep{stab2014identifying} has aimed to identify these argumentative units and classify them into claims, premises, and major claims, or claims describing the key concept in a text. More recently, \citet{chakrabarty2020ampersand} propose to fine-tune BERT \citep{devlin2018bert} for identifying argumentative units and relationships between them within a text and across texts. \citet{lenz2020towards} are the first to propose an end-to-end approach for constructing an \textit{argument graph} \citep{stede2016parallel}, a structured representation of claims and premises in an argumentative text; the graph is built by connecting claim and premise argumentative discourse units. We build on this framework for modeling discourse in conversational data. 
\paragraph{Few-Shot Summarization} As the datasets we introduce are not on a scale with larger datasets, we focus on few-shot and domain transfer summarization techniques. \citet{wang2019exploring} examine domain adaptation in extractive summarization, while \citet{hua-wang-2017-pilot} examine domain adaptation between opinion and news summarization.
Within unsupervised abstractive summarization, several approaches have made use of variational autoencoders \citep{baziotis-etal-2019-seq, chu2019meansum, brazinskas-etal-2020-unsupervised} and pretrained language models \citep{zhou-rush-2019-simple,laban-etal-2020-summary}. \par
Recent work in abstractive \citep{zhang2019pegasus, fabbri2020improving} and extractive-compressive summarization \citep{desai-etal-2020-compressive} has shown the power of pretrained models for a few-shot transfer. The quality of models trained on several hundred examples in these papers is comparable to that of models trained on the equivalent full datasets. Thus, we believe that introducing curated validation and testing datasets consisting of a few hundred examples is a valuable contribution within the current paradigm, which was confirmed by the poor performance of models transferred from other domains compared to that trained on this validation data. 
\DatasetComparison
\section{ConvoSumm}\label{sec:dataset_crowdsourcing}
In this section, we introduce our dataset selection, our annotation protocol, and the characteristics of our crowdsourced dataset. 
\paragraph{Data Selection} For the news comments subdomain, we use the NYT Comments dataset, which consists of 2 million comments made on 9,000  New York Times articles published between 2017 and 2018. It is publicly available and has been used in work for news-comment relevance modeling \cite{kolhatkar2017using}; it also contains metadata that may be of use in summarization modeling. For the discussion forums and debate subdomain, we select Reddit data from CoarseDiscourse \cite{zhang2017characterizing}, which contains annotations about the discourse structure of the threads. For the community question answering subdomain, we use StackExchange (Stack), which provides access to all forums and has been used in modeling for answer relevance and question deduplication \cite{hoogeveen2015cqadupstack}. We chose StackExchange over the commonly-used Yahoo! Answers data due to licensing reasons. For the email threads subdomain, we use the publicly-available W3C corpus \cite{Craswell-2005-TREC}. Previous work also made use of this dataset for email summarization \cite{JanAAAI08} but provided only a small sample of 40 email threads, for which we provide transfer testing results. 
\par
We generally follow the guidance of \citet{tomasoni-huang-2010-metadata}, from summarizing community question answering forums, for determining which subsets of data to select from the above datasets. We remove an example if (1) there were less than five posts (four in the case of email threads; ``post'' refers to any answer, comment, or email); (2) the longest post was over 400 words; (3) the sum of all post lengths was outside of $[100, 1400]$ words (although we extended this maximum length for NYT comments); or (4) the average length of the posts was outside of the $[50, 300]$ words interval. For Stack data, we first filtered answers which received a negative community rating, as defined by the number of user upvotes minus the number of user downvotes. While real-world settings may contain much longer threads, we later show that this setting is already challenging.
\paragraph{Annotation Protocol}
We designed annotation instructions for crowdsourced workers to write abstractive summaries for each of the four datasets, motivated by work in summarizing viewpoints present in online conversation \cite{barker2016summarizing}. We present the crowdsource workers with the data threads, along with any available metadata. For NYT, we presented the workers with the article headline, keywords, and, rather than providing the entire article as context, an extractive BERT-based summary \cite{miller2019leveraging} of the article. We use a BERT summary to give the annotators an idea of the topic of the article. We avoided having annotators read the entire article since the focus of their summaries was solely the content of the comments as per the annotation protocols, and reading the entire article could end up introducing information in the summaries that was not necessarily representative of the comments' main points. We found that these summaries were useful in initial in-house annotations, and allowed us to better understand the context of the comments being summarized. For Reddit and Stack, question tags and information about the subforum were provided; the Stack data includes both answers and answer comments. Reddit data was filtered simply on word limits due to the unavailability of up/down votes from the Coarse Discourse data. Stack data includes the prompt/title as well. Whenever possible, we included username information and the scores of all comments, posts, and answers. 
\DatasetComparisonMDS
\par
Although the instructions differed slightly with the specific nuances of each dataset, they had standard overall rules: (1) summaries should be an analysis of the given input rather than another response or utterance; (2) summaries should be abstractive, i.e., annotators were required to paraphrase and could not repeat more than five words in a row from the source; and (3) summary lengths should contain $[40,90]$ tokens. Following the issues--viewpoints-- assertions framework presented in \citet{barker-gaizauskas-2016-summarizing}, we also instructed annotators that summaries should summarize all viewpoints in the input and should try to include specific details from assertions and anecdotes (unless this made the summary too lengthy). Summarizing based on similar viewpoints is analogous to clustering then summarizing, similar to the comment label grouping procedure before summarization in \citet{barker2016sensei}. To help with this, we recommended wording such as ``Most commenters suggest that...'' and ``Some commenters think that...'' to group responses with similar viewpoints.
\par
However, the email dataset was unique among the selected datasets given that it contained more back-and-forth dialogue than clusters of viewpoints, and thus identifying the speakers was essential to creating summaries that still retained meaning from the original email dialogue. Since the email threads contained fewer individual speakers than the other datasets, this sort of summarization remained feasible. Thus, for this dataset, annotators were instructed to specify the speakers when summarizing the conversation.
\paragraph{Quality-Controlled Crowdsourcing} We crowdsourced our data using Amazon Mechanical Turk. We required that our workers be native English speakers and pass a qualifying exam for each domain to be summarized. We worked with a select group of about 15 workers who formed a community of high-quality annotators. Example summaries were provided to the workers. The workers submitted the qualifying exam, and then one of the authors of this paper provided feedback. If the worker was not sure of the quality of the summaries written, at any point, they could enlist the input of one of the authors.
\par
Additionally, after the workers wrote all summaries, we manually reviewed every summary and made corrections to grammar, wording, and overall structure. Summaries we could not fix ourselves, either because they were poorly written or did not follow the annotation protocols, were flagged to be re-written. They were then sent to our approved group of workers to be re-written, excluding any workers who had written a flagged summary. While data crowdsourced from non-experts may contain noise \cite{gillick2010non}, we believe that our setup of working closely with a small group of workers, providing feedback to individual workers, and manually reviewing all final summaries mitigates these issues.
\ArgumentGraph
\paragraph{Dataset Statistics} We provide statistics in Table \ref{tab:statistics}. The percentage of novel n-grams in our summaries is higher than that of the very abstractive XSum dataset \cite{Narayan2018DontGM} (35.76/83.45/95.50 -\% novel uni/bi/tri-grams). This level of abstraction is likely due to the instructions to perform abstractive summarization and the summaries being an analysis of the input, which results in the insertion of new words (e.g. ``commenters'' likely isn't seen in the input).  The influence of this abstraction is further seen by an analysis of the Extractive Oracle, for which we show ROUGE-1/2/L \cite{lin-2004-rouge}. We see that the performance of an extractive model is above the Extractive Oracle on the very abstractive XSum \cite{Narayan2018DontGM} (29.79 ROUGE-1), but much lower than the Extractive Oracle on the CNN-DailyMail (CNNDM) dataset \cite{nallapati-etal-2016-abstractive} ($>$50 ROUGE-1). The summary lengths are fairly consistent, while the input lengths are the longest for NYT and Stack data. We include the title and additional meta-data such as the headline and snippet in NYT data in input length calculations. 
\par
We analyze multi-document summarization--specific characteristics of our datasets, as proposed by \citet{dey2020corpora}. In particular, inter-document similarity measures the degree of overlap of semantic units in the candidate documents, with scores further from zero signifying less overlap. The notion introduced for redundancy measures the overall distribution of semantic units; the farther the score is from zero, the more uniform semantic units are across the entire input, with the maximum when each unit is present only once. Layout bias measures the similarity of multi-sentential documents with the reference. For more precise definitions, we refer the reader to \citet{dey2020corpora}. We provide results for our data in Table \ref{tab:statistics_mds}. Email data exhibits the most inter-document similarity, which follows the intuition that an email thread consists of a focused discussion typically on a single topic. For redundancy, we see Reddit shows the most uniform distribution of semantic units, perhaps due to Reddit threads' less focused nature compared to the remaining datasets. We do not see a particularly strong layout bias across any parts of the input documents. Our datasets exhibit greater or comparable levels of novel-ngrams compared to multi-document summarization datasets such as MultiNews \cite{fabbri-etal-2019-multi} and 
CQASUMM \cite{chowdhury2018cqasumm}. Our Stack subset has lower inter-document similarity, which presents challenges for models which rely strictly on redundancy in the input, and our datasets generally exhibit less layout bias, when compared to the analysis done in \citet{dey-etal-2020-corpora}.
\paragraph{Comparison to Existing Datasets} Although previous work on conversation summarization, before the introduction of SAMSum \citep{gliwa-etal-2019-samsum}, has largely featured unsupervised or few-shot methods, there exist several datasets with reference summaries. These include SENSEI \citep{barker2016sensei} for news comments, the Argumentative Dialogue Summary Corpus (ADS) \citep{misra-etal-2015-argumentative-dialogue-corpus} for discussion forums, and the BC3 \cite{ulrich2009regression} dataset for email data. However, much of the existing datasets are not wide in scope. For example, SENSEI only covers six topics and the ADS Corpus covers one topic and only has 45 dialogues. Furthermore, they each pertain to one subdomain of conversation. Our dataset avoids these issues by covering four diverse subdomains of conversation and having approximately 500 annotated summaries for each subdomain. Additionally, since neural abstractive summarization baselines do not exist for these datasets, we benchmark our models on these datasets to further their use as test sets. We similarly include the AMI and ICSI meeting datasets within our benchmark. 
\par
Within community question answering, the WikiHowQA dataset \citep{deng2019wikihowQA-joint-learning} consists of user response threads to non-factoid questions starting with ``how to," including labels for the answer selection task and reference summaries. The CQASUMM dataset \citep{chowdhury2018cqasumm} sampled threads from Yahoo! Answers in which the best answer could be used as a reference summary. However, this heuristic is not guaranteed to cover all the user answers' perspectives, so we believe our dataset is a more principled benchmark for community question answering.
\par
It is also noted that several large-scale MDS datasets have been introduced in the news domain \citep{fabbri-etal-2019-multi, gu2020generating, ghalandari2020large}, for creating Wikipedia lead-paragraphs \citep{46594}, and for long-form question answering \citep{fan-etal-2019-eli5}. However, these do not focus on the conversational domain.
\section{Argument Graph Summarization}\label{sec:methods}
As our annotation protocol is motivated by the issues-viewpoints-assertions framework proposed in \citet{barker2016summarizing}, we propose to instantiate a modified version of that work's theoretical, proposed graph model.
\paragraph{Argument Graph Construction} We build on the argument graph formulation of \citet{lenz2020towards}, a variant of Argument Interchange Format \cite{chesnevar2006towards}. Claims and premises are represented as information nodes ($I$-nodes), with the relations between them represented as scheme nodes ($S$-nodes). Let $V = I \cup S$ be the set of nodes, and $E \subset V \times V$ the set of edges describing support relationships among the nodes. We then define the argument graph $G = (V, E)$.
\par
\citet{lenz2020towards} breaks the construction of the argument graph down into four steps: (1) \textit{argument extraction}, or the identification of argumentative discourse units; (2) \textit{relationship type classification}, or the classification of edges between nodes; (3) \textit{major claim detection}; and (4) \textit{graph construction}, or the construction of the final graph based on the identified nodes and edges. To adapt this formulation to our multi-document setting, we first perform \textit{argument extraction} and \textit{relationship type classification} for each individual input document and finally \textit{graph construction} to determine relationships among claims from all documents. 
\paragraph{Argument Extraction} 
For extracting arguments from a single document, we build on work in argument mining with pretrained models \cite{chakrabarty2020ampersand}. As in \citet{lenz2020towards}, our argumentative units are sentences, from which we identify \textit{claims}, which are assertions that something is true, and \textit{premises}, which are propositions from which a conclusion is drawn. Additionally, we identify and remove non-argumentative units. We train a three-way classifier for the task of argument extraction, following \citet{chakrabarty2020ampersand} and making use of data for argument mining from that paper and from \citet{stab2014identifying}. The output of this step can also simply be used without further graph construction as a less noisy version of the input, which we call \textbf{-arg-filtered}.
\paragraph{Relationship Type Classification} We follow the procedure in \citet{lenz2020towards} and use entailment to determine the relationship between argumentative units within a document. However, rather than using the classifier provided, we make use of RoBERTa \cite{liu2019roberta} fine-tuned on the MNLI entailment dataset \cite{N18-1101}. Rather than using both support and contradiction edges between claims and premises, we make the simplification that all relationships can be captured with support edges, as we are dealing with a single document in this step. Within a single text, the premise can be tied as following from one of the claims. We create an edge between any premise and the claim it most entails if the entailment score from RoBERTa is greater than 0.33, based on manual analysis of the scores. If a premise is not labeled as supporting a claim, then we heuristically create an edge between that premise and the closest claim preceding it in the text.
\par
Since not all texts in the benchmark datasets may be argumentative or may be too short to contain major claims, we use some heuristics in our graph creation. If none of the argumentative sentences are labeled as claims (i.e., all are labeled as premises) in argument extraction, the text's first sentence is labeled as the claim. Furthermore, we do not identify a single claim as the major claim since there may be multiple major points of discussion.
\BaselineResults
\paragraph{Graph Construction} For the final graph, for each of the documents in an example, we run the above procedure and obtain a set of claims and associated premises. We then identify support edges between claims, which may be across documents. One claim may make a larger assertion, which is supported by other claims. 
We run our entailment model over all potential edges (in both directions) among claims in the document and greedily add edges according to the entailment support score while no cycles are made. After this step, we are left with a set of claims which do not entail any other nodes or, stated otherwise, do not have parent nodes. Following the terminology of \citet{barker-gaizauskas-2016-summarizing}, these nodes can be considered viewpoints. 
\par
We then identify issues or topics on which the viewpoints differ. We run our entailment model for all parent claim nodes again in both directions over these claims and identify nodes that contradict each other with probability over 0.33, based on manual analysis of the resulting graphs. We greedily add edges to maintain a tree structure, joining these nodes to a special node, which we call the Issue node. All Issue nodes, as well as claims which are not connected to any Issue node, are connected to a dummy `Conversation Node' which serves as the root of the argument graph. We show an example Issue subgraph for NYT data in Figure \ref{fig:argument_graph}.
\paragraph{Argument Graphs to Summaries} Recent work has shown the strength of text-based pretrained models on graph-to-text problems \cite{ribeiro2020investigating}. Following that work, we linearize the graph by following a depth-first approach starting from the Conversation Node. We found that inserting special tokens to signify edge types did not improve performance, likely due to the size of our data, and simply make use of an arrow $\to$ to signify the relationship between sentences. We train a sequence-to-sequence model on our linearized graph input,  which we call \textbf{-arg-graph}.
\BARTTransferResults
\section{Experimental Settings}\label{sec:experimental_settings}
We use the fairseq codebase \cite{ott-etal-2019-fairseq} for our experiments. Our base abstractive text summarization model is BART-large \cite{lewis-etal-2020-bart}, a pretrained denoising autoencoder with 336M parameters that builds on the sequence-to-sequence transformer of \citet{vaswani2017attention}. We fine-tune BART using a polynomial decay learning rate scheduler with Adam optimizer \citep{kingma2014method}. We used a learning rate of  3e-5 and warmup and total updates of 20 and 200, following previous few-shot transfer work \cite{fabbri2020improving}. We could have equally fine-tuned other pretrained models such as Pegasus \cite{zhang2019pegasus} or T5 \cite{raffel2019exploring}, but \citet{fabbri2020improving} find that BART largely performs equally well in few-shot settings when compared to Pegasus.
\par
For the NYT and Stack datasets, which contain sequences over the typical 1024 max encoder length with which BART is trained, we copied the encoder positional embeddings to allow sequences up to length 2048. To address the input-length of meeting summaries, which range from 6k to 12k tokens, we use the Longformer \cite{Beltagy2020Longformer}, which allows for sequences up to length 16k tokens. We initialize the Longformer model with BART parameters trained on the CNN-DailyMail dataset, as the meeting summarization datasets contain fewer than 100 data points. We otherwise fine-tune models from vanilla BART, following intuition in few-shot summarization \cite{fabbri2020improving} and based on initial experiments. In the tables which follow,  ''-arg" refers to any model trained with argument-mining-based input, and we specify which -arg-graph or -arg-filtered settings were used for each dataset below.
%
%

\BARTMeetingResults
\section{Results}\label{sec:results}
\par
We provide results for baseline, unsupervised extractive models in Table \ref{tab:extractive_baselines}. Lexrank \cite{erkan2004lexrank} and Textrank \cite{mihalcea2004textrank}, and BERT-ext \cite{miller2019leveraging}, which makes use of BERT \cite{devlin2018bert}. The unsupervised extractive models perform well below the extractive oracle performance, suggesting the difficulty of content selection in this setting. 
\par
We train BART on 200 examples from our validation set for abstractive models, using the remaining 50 as validation and test on the final test set of 250 examples. We tested zero-shot transfer from CNNDM and SAMSum in zero-shot settings, although these resulted in a much lower performance of about 28 ROUGE-1. Few-shot model performance is shown in Table \ref{tab:bart_transfer}. The abstractive model performs at or above the Extractive Oracle, suggesting the need for better abstractive models.
\par
We also train on our argument mining-based approaches and show results in Table \ref{tab:bart_transfer}. We see ROUGE improvements when applying BART-arg-graph for Reddit, and Stack data. The -arg-filtered variation (which, as defined in Section 4, is the less noisy version of the input produced by the argument extraction step) outperformed the -arg-graph variation on both email and NYT data. For email data, however, this did not improve upon the BART baseline, likely due to the dataset's characteristics; email data is shorter and more linear, not benefiting from modeling the argument structure or removing non-argumentative units. We provide full results for both variations in the Appendix. 
\paragraph{Benchmarking Other Conversation Summarization Datasets} We benchmark our models on widely used meeting summarization datasets. Due to the input's linear nature and the size of the meeting transcripts, we found improved results using -arg-filtered to filter non-argumentative units rather than incorporating the graph structure. Results are shown in Table \ref{tab:meeting}. The Longformer model performs as well or better than previous state-of-the-art results on these datasets, despite not making use of more complex modeling structures, and we generally see improvement with argument-mining. 
\par
As noted above, there exist prior datasets for dialogue, community question answering, email, forum, and news comments summarization. We benchmark results on these datasets in Table \ref{tab:benchmark_others}. We outperform prior work on SAMSum \cite{gliwa-etal-2019-samsum}, and CQASUMM \cite{chowdhury2018cqasumm} with our BART and BART-arg-graph models, respectively. We did not find improvement on SAMSum with the BART-arg model due to the extremely short and focused nature of the dialogues, analogous to email data performance. We also provide transfer results of BART and BART-arg-graph models from our email and news-comment data to BC3 \cite{ulrich2009regression}, ADS \cite{misra-etal-2015-argumentative-dialogue-corpus}, and SENSEI data \cite{barker2016sensei}, for which no prior neural abstractive summarization results existed. 
\BARTBenchmarkOthers
\paragraph{Human Evaluations} We collect human judgment annotations for two of the four quality dimensions studied in \citet{kryscinski-etal-2019-neural} and \citet{fabbri2020summeval}, namely consistency and relevance. Consistency is defined as the factual alignment between the summary and the summarized source text, while relevance is defined as the summary's ability to select important content; only relevant information and viewpoints should be included. We did not include fluency as an initial inspection of the data found fluency to be of very high quality, as has shown to be the case for pretrained models in news summarization \cite{fabbri2020summeval}. We did not include coherence as this was generally not an issue of concern in the initial analysis. 
\par
We randomly select 25 random examples from the Reddit corpus and ten examples from the AMI corpus, and output from the BART and BART-arg-graph models. These data points were chosen to demonstrate what characteristics are realized in differences across ROUGE for argument-graph and argument-noise-reduction approaches. Ten examples were chosen from AMI due to the size of the input and annotation constraints.  The annotator sees the source article and randomly-ordered output from the model and then rates the summaries for relevance and consistency on a Likert from 1 to 5, with 5 being the best score. We averaged the score of three native English-speaking annotators on each example and then across examples. Results are shown in Table  \ref{tab:table_human_evaluation}. We find that the annotators prefer our argument mining-based approaches in both dimensions. However, the results are close. Furthermore, the scores for relevance and consistency are rather low, especially on the Reddit dataset and when compared to results on the CNN-DailyMail Dataset from \citet{fabbri2020summeval}. These results demonstrate the difficulty of modeling such conversational data. Examples are included in the appendix. 
\TableHumanEvaluation
\section{Conclusion}\label{sec:conclusion}
We propose ConvoSumm, a benchmark of four new, crowdsourced conversation datasets and state-of-the-art baselines on widely-used datasets that promote more unified progress in summarization beyond the news domain. Our benchmark consists of high-quality, human-written summaries that call for abstractive summaries and a deeper understanding of the input texts' structure. We provide results for baseline models and propose to model the text's argument structure, showing that such structure helps better quantify viewpoints in non-linear input in both automatic and human evaluations. Our analysis notes challenges in modeling relevance and consistency in abstractive conversation summarization when compared to news summarization. 
\section{Ethical Considerations}\label{sec:ethical_considerations}
As we propose novel conversation summarization datasets and modeling components, this section is divided into the following two parts.
\subsection{New Dataset}
\paragraph{Intellectual Properties and Privacy Rights}
All data for our newly-introduced datasets are available online; please see the following for New York Times comment data\footnote{\url{https://www.kaggle.com/aashita/nyt-comments}}, StackExchange data\footnote{\url{https://archive.org/download/stackexchange}}, and W3C email data\footnote{\url{https://tides.umiacs.umd.edu/webtrec/trecent/parsed_w3c_corpus.html}}. Reddit data is available via the Google BigQuery tool\footnote{\url{https://console.cloud.google.com/bigquery}}.
\paragraph{Compensation for Annotators}
We compensated the Turkers approximately \$12--\$15 per hour. We first annotated examples in-house to determine the required annotation speed. Typically, the summarization task took around 10 minutes, and we compensated the workers from \$2.25 to \$3.00 per task, depending on the domain and deadline requirements. 
\paragraph{Steps Taken to Avoid Potential Problems}
We interacted closely with the Turkers to ensure that compensation was fair and that the instructions were clear. To maintain the quality of the dataset, we manually reviewed the crowdsourced summaries for language use. Initial investigation into Reddit data showed certain inappropriate language usage, so we filtered these examples automatically. 
\subsection{NLP Application}
\paragraph{Bias}
Biases may exist in the datasets, such as political bias in the news datasets and gender bias in potentially all of the datasets. Thus, models trained on these datasets may propagate these biases. We removed data with offensive language when possible. 
\paragraph{Misuse Potential and Failure Mode}
When used as intended, applying the summarization models described in this paper can save people much time. However, the current models are still prone to producing hallucinated summaries, and in such a case, they may contribute to misinformation on the internet. Further research is needed to ensure the faithfulness of abstractive summaries to address this issue, as this issue is present among all current abstractive summarization models. 
\paragraph{Environmental Cost}
The experiments described in the paper make use of V100 GPUs. We used up to 8 GPUs per experiment (depending on the experiment; sometimes, a single GPU was used to run the maximum number of experiments in parallel). The experiments may take up to a couple of hours for the larger datasets. Several dozen experiments were run due to parameter search, and future work should experiment with distilled models for more light-weight training. We note that while our work required extensive experiments to draw sound conclusions, future work will be able to draw on these insights and need not run as many large-scale comparisons. Models in production may be trained once for use using the most promising settings.  


\bibliography{anthology,acl2021}
\bibliographystyle{acl_natbib}
\appendix
\section{Full Results}
We present the results of BART and -arg variations on our four crowdsourced datasets in Table \ref{tab:bart_transfer_full}.
\BARTTransferFullResults
\section{Sample Output}
We provide examples of model outputs to offer more insight into the datasets and models. An example of Reddit input and outputs for which the models remain faithful to the source is found in Table \ref{tab:example_summaries_first}. The gold summary balances being a meta-analysis of the input documents with providing sufficient details. We provide an additional example of outputs that struggle with consistency and relevance in Table \ref{tab:example_summaries_second}. In the BART output, the model mistakes the suggestion in the input to pay debt before starting a business. In BART-arg, the model incorrectly determines relevance, as the suggestion that one should invest in pumpkins was sarcastic and not emphasized in the input. This output points to a need to better model interactions and salience in the conversation data. 
\section{Additional Details}
For reproducibility purposes, we provide (smallest NLL loss, largest NLL loss) bounds on the validation loss from non-baseline models encountered when training on our newly-proposed datasets. NYT (5.77, 5.83); Reddit: (5.62, 5.69); Stack (5.60, 5.63); Email: (5.40, 5.45).
\ExampleSummariesFirst
\ExampleSummariesSecond

\end{document}